\documentclass[letterpaper, 10 pt, conference]{ieeeconf}  

\IEEEoverridecommandlockouts                              

\usepackage{amsmath} 
\usepackage{amssymb}  

\title{\LARGE \bf
MambaControl: Anatomy Graph-Enhanced Mamba ControlNet with \\Fourier Refinement for  Diffusion-Based Disease Trajectory Prediction
}

\author{Hao Yang$^{1}$, Tao Tan$^{1}$, Shuai Tan$^{2}$, Weiqin Yang$^{3}$, Kunyan Cai$^{1}$, Calvin Chen$^{4}$, Yue Sun$^{1}$$^{,}$$^{\ast}$
\thanks{$^{\ast}$ Corresponding author: Yue Sun. Contact: {\tt\small yuesun@mpu.edu.mo}}
\thanks{$^{1}$Hao Yang, Tao Tan, Kunyan Cai and Yue Sun  are with the Faculty of Applied Sciences, Macao Polytechnic University, Rua de Luis Gonzaga Gomes, Macao, China.}
\thanks{$^{2}$Shuai Tan is with Department of  Electrical Engineering, Zhejiang University, China.}%
\thanks{$^{3}$Weiqin Yang with  Department of  Computer Science,  The University of Adelaide, Australia.}%
\thanks{$^{4}$Calvin Chen with  Department of  Computer Science,  University of Birmingham, UK.}
}

\usepackage{multirow} 
\usepackage{booktabs}
\usepackage[table]{xcolor}

\usepackage{graphicx}
\usepackage{amsmath}
\usepackage{amssymb}
\usepackage{booktabs}
\usepackage{pifont}

\usepackage{algorithmicx}
\usepackage{algorithm}
\usepackage{algpseudocode}

\usepackage{cleveref}

\newcommand{\ourmodel}[0]{MambaControl}

\begin{document}

\maketitle
\thispagestyle{empty}
\pagestyle{empty}

\begin{abstract}
Modelling disease progression in precision medicine requires capturing complex spatio-temporal dynamics while preserving anatomical integrity. Existing methods often struggle with longitudinal dependencies and structural consistency  in progressive disorders. To address these limitations, we introduce \ourmodel, a novel framework that integrates selective state-space modelling with diffusion processes for high-fidelity prediction of medical image trajectories. To better capture subtle structural changes over time while maintaining anatomical consistency, MambaControl combines Mamba-based long-range modelling with graph-guided anatomical control to more effectively represent anatomical correlations. Furthermore, we introduce Fourier-enhanced spectral graph representations to capture spatial coherence and multiscale detail, enabling MambaControl to achieve state-of-the-art performance in Alzheimer's disease prediction. Quantitative and regional evaluations demonstrate improved progression prediction quality and anatomical fidelity, highlighting its potential for personalised prognosis and clinical decision support.

\end{abstract}

\section{INTRODUCTION}
Modeling disease progression in medical imaging represents a critical challenge in computational medicine, with profound implications for diagnosis, prognosis, and therapeutic development~\cite{ahmed2020artificial}. Accurate prediction of how diseases evolve over time at the individual level, particularly through the analysis of medical images, is essential for personalized medicine~\cite{mohsin2023role}. Such capabilities enable personalized interventions and more efficient clinical trials~\cite{young2024data}. This task is especially challenging when dealing with complex heterogeneous conditions, such as Alzheimer's disease (AD), where subtle structural changes such as progressive brain atrophy often manifest gradually over extended periods, making accurate modeling vitally important~\cite{weiner2013alzheimer}. 

   \begin{figure}[thpb]
      \centering
      \includegraphics[scale=0.72]{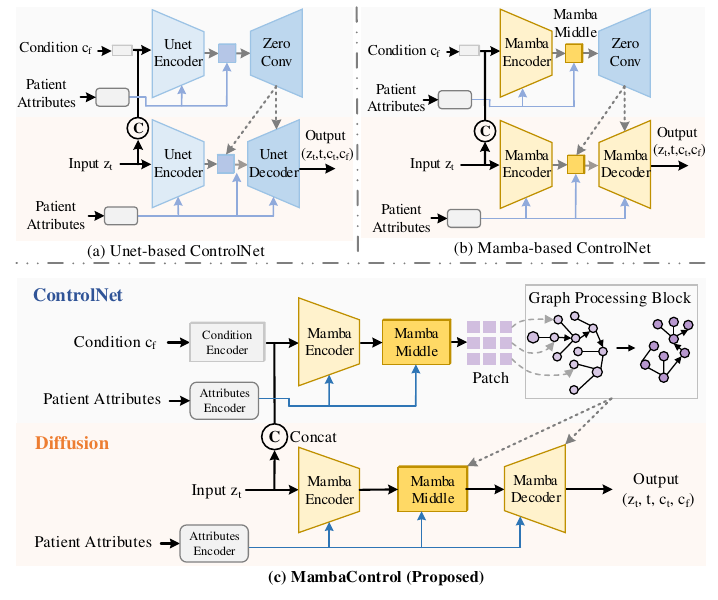}
      \caption{Comparison of ControlNet architectures for guiding diffusion models. (a) shows the baseline U-Net-based ControlNet. (b) shows a direct replacement using Mamba blocks. (c) shows our proposed MambaControl method, which utilizes separate Mamba pathways for control and diffusion, embedding graph-processed anatomical features from the control pathway to guide the main diffusion pathway generation.}
      \label{fig:compar}
   \end{figure}
Traditional approaches for modeling disease progression have relied mainly on statistical analysis of scalar biomarkers~\cite{oxtoby2017imaging,young2024data}, which fail to capture the rich spatial information embedded within medical images. These conventional methods typically construct population-level trajectories, providing limited insight into patient-specific disease dynamics~\cite{marinescu2018tadpole}. Although advanced neuroimaging techniques now provide the necessary longitudinal data with rich spatial detail~\cite{lorenzi2019probabilistic}, effectively leveraging this information remains computationally challenging. Key difficulties include capturing long-range temporal dependencies between image sequences and maintaining spatial coherence, anatomical plausibility, and fidelity within the high-dimensional predictions for each individual.



Generative deep learning architectures, notably Variational Autoencoders (VAEs)~\cite{sauty2022progression}, Generative Adversarial Networks (GANs)~\cite{xia2021learning}, and diffusion models~\cite{pinaya2022brain,puglisi2024enhancing}, have demonstrated potential for synthesizing realistic medical images that depict disease states across different time points. However, existing methods still face significant limitations in addressing the challenges mentioned above. Some approaches such as CounterSynth~\cite{pombo2023equitable} struggle to effectively incorporate and leverage the valuable temporal information present in longitudinal data when available. While methods such as SADM~\cite{yoon2023sadm} attempt longitudinal modeling, they can suffer from high computational demands and may lack robust mechanisms for conditioning predictions on patient-specific attributes or ensuring strict anatomical consistency. Furthermore, even recent diffusion-based models such as BrLP~\cite{puglisi2024enhancing}, while integrating prior knowledge, often rely on standard UNet backbones. The convolutional nature of UNets inherently limits their capacity to model complex, long-range temporal relationships across extended image sequences and can struggle to enforce fine-grained structural constraints derived from longitudinal information.

Thus, a critical challenge remains in developing frameworks that can simultaneously and effectively model both long-range temporal dynamics and intricate spatial details with high fidelity and anatomical coherence. Addressing this critical challenge, we introduce MambaControl, a novel framework for disease progression modeling (conceptually illustrated in Figure \ref{fig:compar}(c)). Specifically designed to enhance both temporal dependency modeling and spatial-anatomical coherence, our approach integrates state space models (SSMs) within a conditional diffusion framework guided by graph-based structural priors. Our contributions can be summarized as follows:

\begin{itemize}

\item \textbf{Enhanced Temporal Dependency Modeling:} We leverage Mamba's selective state space modeling to replace the conventional UNet architecture in diffusion models, significantly improving the model’s ability to capture long-range temporal dependencies in disease progression data.
\item \textbf{Coherent Longitudinal Data Integration:} We incorporate the Mamba architecture as ControlNet, using topological graph priors to explicitly model longitudinal patient data. This ensures that the generated predictions adhere to pathophysiologically valid structures while fully leveraging each subject’s complete temporal history.
\item \textbf{Improved Fidelity and Coherence:} We enhance multi-scale representations using Fourier-based graph networks, refining the model's ability to maintain temporal coherence and anatomical fidelity, and enabling smooth, realistic structural changes in predicted images over time.

\end{itemize}


\section{related work}
\subsection{Deep Learning for Image-base Disease Progression}

Deep learning models have been increasingly explored for modeling disease progression from medical images. VAEs offered early promise by learning compact latent representations for predicting future brain MRIs~\cite{basu2019early}. Subsequent longitudinal VAEs incorporated interpretable parameters and allowed image generation at arbitrary disease stages~\cite{sauty2022progression}. However, VAEs often face challenges in generating high-fidelity images with fine anatomical details.

GANs have shown promise in synthesizing realistic medical images. While advancements include techniques to enhance output coherence and transfer learning for individualization~\cite{xia2021learning}, GANs still contend with training instability and difficulties in fully capturing complex anatomical variability [10].

More recently, Diffusion models gained prominence due to their stable training and high-quality image production~\cite{rombach2022high}.
Pinaya et al.~\cite{pinaya2022brain} developed a latent diffusion model for brain MRI generation customized by conditions such as age and sex, but focused only on single time point generation. Yoon et al.~\cite{yoon2023sadm} introduced SADM, specifically designed for longitudinal medical image generation, enhancing image coherence through explicit modeling of temporal dependencies. Nevertheless, SADM lacks mechanisms to incorporate disease-specific prior knowledge, failing to leverage known disease progression patterns to enhance generation quality. To address these limitations, Puglisi et al.~\cite{puglisi2024enhancing} proposed BrLP, a latent diffusion model for brain disease progression that incorporates prior knowledge from disease models. While these approaches have demonstrated significant advancements, they typically rely on UNet architectures within their diffusion frameworks, which limit their capacity to model complex temporal dependencies efficiently.

\subsection{State Space Models for Sequential Data Analysis}

Traditional SSMs are built upon linear time-invariant systems but have been adapted for deep learning through various parameterizations and training techniques~\cite{behrouz2024chimera}. A key development is Mamba~\cite{gu2023mamba}, a selective SSM building upon prior work such as S4~\cite{gu2021efficiently}. Mamba employs a data-dependent selection mechanism and exhibits linear computational complexity with respect to sequence length, offering significant advantages over attention-based transformers for processing long sequences. Its selective design is particularly well-suited for modeling processes with multiple temporal scales, such as disease progression.

The application of SSMs in medical imaging is emerging. Studies have shown their effectiveness in temporal modeling for clinical outcome prediction~\cite{cauchi2024individualized}, deformable image registration~\cite{guo2024mambamorph}, and both 2D~\cite{ma2024u} and 3D~\cite{liu2024mambav3d} medical image segmentation, often achieving competitive performance with reduced computational cost compared to transformer-based methods. However, these applications predominantly address single-timepoint tasks.

\section{Methodology}

\begin{figure*}
    \centering
    \includegraphics[width=1\linewidth]{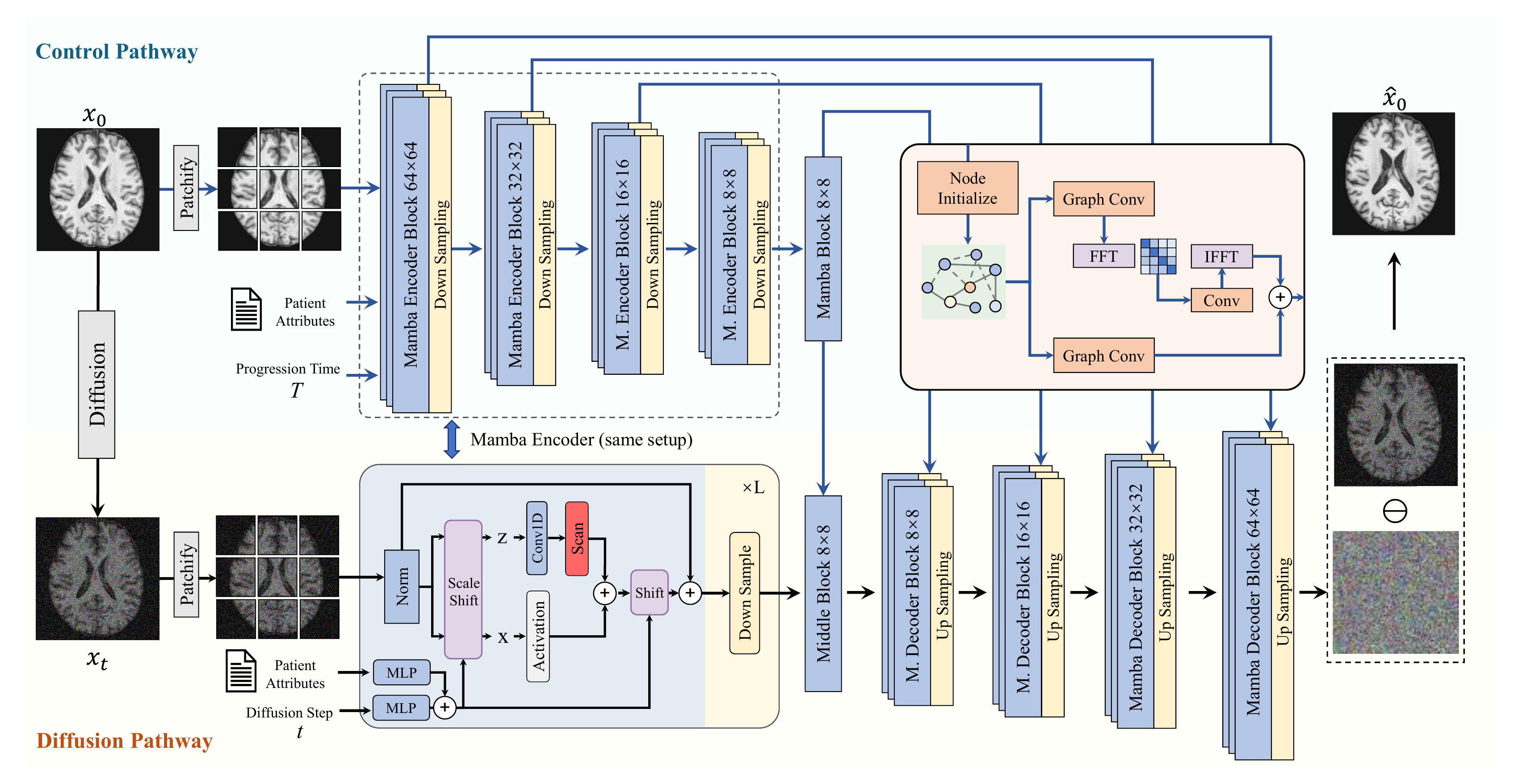}
    \caption{Framework of our proposed \ourmodel. It consists of two Mamba-based pathways: Diffusion and Control. The Control pathway incorporates a graph convolutional network in its downsampling phase to extract anatomical and spatial features. These features are then integrated into the Diffusion pathway's upsampling stage, enabling precise control over the generation process.}
    \label{fig:arch}
\end{figure*}

We present MambaControl (Figure \ref{fig:arch}) to tackle the challenge of simultaneously modeling the long-term temporal dynamics and intricate spatial-structural changes characteristic of disease progression. MambaControl is architected around three interconnected modules. It employs a Mamba-based diffusion model (Section $ \ref{sec:dim} $) for efficient sequence modeling over extended periods, crucial for tracking disease trajectories. Complementing this, a Fourier-enhanced graph module (Section $ \ref{sec:graph} $) encodes detailed spatial relationships and structural information derived from anatomical regions. Synergy between temporal dynamics and spatial structure is achieved through our core MambaControl mechanism (Section $ \ref{sec:control} $), which integrates the graph representation to guide the Mamba-driven diffusion process.

\subsection{Diffusion Mamba}
\label{sec:dim}
Mamba has emerged as a structured SSM~\cite{ssm_s4} capable of efficient sequence modeling, which can be leveraged to refine the process of diffusion, particularly in conditioning mechanisms. In a continuous-time SSM, an input signal $x(i)$ is first encoded into a hidden state vector $h(i)$ and then decoded into the output signal ${y}(i)$ according to the following Ordinary Differential Equations (ODEs):
\begin{align}
    {h}'(i) &= \mathbf{A}{h}(i) + \mathbf{B}x(i), \\
    {y}(i) &=  \mathbf{C}{h}'(i) + \mathbf{D}x(i),
\end{align}
where ${h}'$ denotes the derivative of $h$, and $\mathbf{A}$, $\mathbf{B}$, $\mathbf{C}$, $\mathbf{D}$ denote the weights of the SSM. To learn this process in  two-dimensional vision as a discrete signal, it  Zero-Order Hold (ZOH) rule for discretization, which the above ODEs can be recurrently solved:
\begin{align}
    \mathbf{\bar{A}} &= \exp \left( \mathbf{\Delta} \mathbf{A} \right), \\
    \mathbf{\bar{B}} &= \left( \mathbf{\Delta} \mathbf{A} \right)^{-1} \left( \exp \left( \mathbf{\Delta} \mathbf{A} \right) -\mathbf{I} \right) \cdot \mathbf{\Delta} \mathbf{B}, \\
    {h}_{i} &= \mathbf{\bar{A}}{h}_{i-1} + \mathbf{\bar{B}}{x}_{i}, \\
    {y}_{i} &= \mathbf{C}{h}_{i} + \mathbf{D}{x}_{i},
\end{align}
where $\mathbf{\Delta}$ is the fixed time step term. The matrixs in Mamba~\cite{mamba} are dynamic weights dependant on the input $x$. 

Building on Mamba model analysis, we now introduce the diffusion model and its integration methodology.  Diffusion models generate data through progressive noise addition and removal processes. Consider the standard diffusion process described by the forward stochastic differential equation (SDE):
\begin{equation}
    \text d\mathbf{x}_t = f(t)\mathbf{x}_t \text dt + g(t) \text d\mathbf{w}_t,
\end{equation}
where $\mathbf{x}_t \in \mathbb{R}^d$ represents the state of the data at time $t$, $f(t)$ and $g(t)$ are drift and diffusion coefficients respectively, and $\mathbf{w}_t$ is a standard Wiener process. The generative process is given by the reverse-time SDE:
\begin{equation}
    \text d\mathbf{x}_t = \left[f(t)\mathbf{x}_t - g^2(t) \nabla_{\mathbf{x}_t} \log p_t(\mathbf{x}_t) \right]\text dt + g(t) \text d\bar{\mathbf{w}}_t,
\end{equation}
where $p_t(\mathbf{x}_t)$ denotes the marginal density. The primary challenge lies in estimating the score function $ \nabla_{\mathbf{x}_t} \log p_t(\mathbf{x}_t) $, which is learned by training a Mamba model to parameterise the score network, expressed as $ \epsilon_\theta(\mathbf{x}_t, t) = \text{Mamba}(\mathbf{x}_t, t) $, to approximate it. In the standard diffusion framework, the forward noise process is characterized by: 
\begin{equation}
    \mathbf{x}_t = \sqrt{\alpha_t} \mathbf{x}_0 + \sqrt{1 - \alpha_t} \boldsymbol{\epsilon}, \quad \boldsymbol{\epsilon} \sim \mathcal{N}(0, \mathbf{I}).
\end{equation}

We propose integrating Mamba into the diffusion denoising function $\epsilon_\theta$ as follows:
\begin{equation}
    \epsilon_\theta(\mathbf{x}_t, t, \mathbf{c}) = \text{Mamba}(\mathbf{x}_t, t, \mathbf{c}).
\end{equation}

\subsection{Fourier-enhanced Anatonomy Graph}
\label{sec:graph}

To better factorize the feature in a patient anatomy-wise behavior, we propose a graph-based structural representation for modeling anatomical regions and their relationships. Our method comprises three interconnected components that enable effective anatomical modeling and feature extraction.

First, we consider an input image $\mathbf{x} \in \mathbb{R}^{H \times W \times C}$, which is divided into non-overlapping patches $\{ \mathbf{p}_i \}_{i=1}^{N}$, where $N$ is the number of patches. Each patch is represented as a node in a graph $G = (V, E)$, where $V$ represents the set of patches and $E$ represents the connectivity between patches, determined via a spatial proximity function or learned graph adjacency. Each patch is encoded using the Mamba SSM:
\begin{equation}
\mathbf{h}_t = \sigma(W_1 \mathbf{x}_t) \cdot (\mathbf A \mathbf{h}_{t-1} + \mathbf  B \mathbf{x}_t) + W_2 \mathbf{x}_t,
\end{equation}
where $\mathbf{A}$, $\mathbf{B}$ are learned state-space matrices, and $W_1, W_2$ are projection matrices. The encoded features form the graph node representations.

Next, the Mamba features are aggregated as $\mathbf{H} = [\mathbf{h}_1, \mathbf{h}_2, \dots, \mathbf{h}_N]^\top \in \mathbb{R}^{N \times d}$, defining a graph signal on $G$. The graph structure is captured by the normalised Laplacian:

\begin{equation}
L = I - D^{-\frac{1}{2}} A_d D^{-\frac{1}{2}}, {A}_d = \text{Softmax}(\mathbf{H}_t \mathbf{H}_t^\top),
\end{equation}
where $A_d$ is the adjacency matrix and $D$ is the degree matrix. With the graph structure established, we further focus on information exchange between anatomical regions. We propagate information across patches using the graph Laplacian. Graph spatial convolution updates node features as $\mathbf{H}' = \sigma( \tilde{L} \mathbf{H} W_g )$, with $W_g$ as a trainable weight matrix, $\sigma(\cdot)$ is the sigmoid function. 

Then, to enforce Fourier component learning, given a graph with Laplacian matrix $L$, we perform spectral decomposition of $U \Lambda U^\top= L $, where $U$ contains eigenvectors and $\Lambda = \text{diag}(\lambda_1, \lambda_2, ..., \lambda_N)$ holds the eigenvalues, representing frequency components. 
Spectral filtering modifies frequency components through a function $g(\Lambda)$:
\begin{equation}
    \quad g(\Lambda) = \sum_{k=0}^{K} \theta_k T_k(\Lambda),
\end{equation}
where $T_k(\Lambda)$ are Chebyshev polynomials for efficient approximation and $\theta_k$ are series of learnable parameter. Applying spectral filtering in the node space yields the final representation:
\begin{equation}
    \mathbf{H}' = \sigma(U g(\Lambda) U^\top \mathbf{H} W_g), 
\end{equation} which can be represented by $\mathbf{H}' = G_t(\mathbf{H})$, where $\mathbf{H} = \text{Mamba}_G(\mathbf x_t, t)$, $M$ represent a mamba model  predicts new patch dependencies as the diffusion process evolves. Graph representation enhances context modeling, representing patches as graph nodes allows for global interactions and preserves spatial relationships. Using Fourier-enhanced embeddings for patches ensures that fine-grained structures are captured effectively.


\subsection{Integration of Graph-based Control with \ourmodel} \label{sec:control}

The preceding sections have introduced \ourmodel, which leverages a Fourier-enhanced anatomical graph representation. We now present the integration of these components within a unified framework for disease progression prediction.

ControlNet extends diffusion models by incorporating additional control conditions that guide the generative process at each time step. This is formally expressed as
\begin{equation} 
\mathbf{x}_{t+1} = \mathbf{x}_t + \epsilon_\theta(\mathbf{x}_t, t, \mathbf{c}), \end{equation} 
where $\mathbf{c}$ denotes auxiliary control signals such as depth maps, segmentation masks, or edge maps. In the present study, we replace the conventional UNet architecture in ControlNet with Mamba, a model capable of memory-efficient, long-range sequence learning. This substitution is particularly well suited to medical imaging tasks, where modelling temporal dependencies and detecting subtle anatomical variations are critical for understanding disease evolution.

To enhance the controllability of the diffusion process, we incorporate the structured guidance of MambaControl directly into the generative mechanism. Let $\epsilon_\theta(\mathbf{x}_t, t, \mathbf{c})$ be the denoising function conditioned on control signal $\mathbf{c}$. We redefine this conditioning by embedding the Mamba-encoded graph representation $G_t$ as follows: 
\begin{equation} \hat{\mathbf{x}}_0 = \epsilon_\theta(\mathbf{x}_t, t, G_t (\mathbf{x}_t)). \end{equation}

To maintain the generative integrity of the diffusion process, the graph-based representation $G$ is integrated exclusively within the decoder module of the model. This design aligns with the MambaControl architecture. Notably, the associated weight matrix $W_g$ is initialised with zeros, ensuring that the early stages of generation adhere to a standard denoising trajectory. As the process unfolds, the structured graph information is incrementally introduced, enabling a gradual and stable incorporation of anatomical guidance.

\begin{table*}[t]
    \caption{Results from comparison and ablation study. MAE (± SD) in predicted volumes is expressed as a percentage of total brain volume, while PSNR (± SD) and SSIM (± SD) are used to assess image quality.}
    \label{tab:performance}
    \setlength{\tabcolsep}{5pt}
    \def\arraystretch{1.5} 
    \resizebox{2.0\columnwidth}{!}{%

    \begin{tabular}{c|l|c|cc|ccccc} \hline 
    &&Ablation. & \multicolumn{2}{c|}{\textbf{Image-based Metrics}} & \multicolumn{5}{c}{\textbf{MAEs of Region Volumes}}  \\
    &\textbf{Method}& (Graph) & PSNR(dB) $\uparrow$ & SSIM(\%) $\uparrow$ & Hippocampus $\downarrow$ & Amygdala $\downarrow$ & Lat. Ventricle $\downarrow$ & Thalamus $\downarrow$ & CSF $\downarrow$ \\
    \hline
    \multirow{9}{*}{\rotatebox[origin=c]{90}{\parbox[c]{4cm}{\centering \textbf{Comparison Study}}}}
    & \multicolumn{9}{l}{\textbf{Single-image (Cross-sectional)}} \\
    \cline{2-10}
    &CardiacAging~\cite{campello2022cardiac} &  - & 27.78 ±  1.49  &  92.04 ±  0.99 & 0.032 ± 0.025 & {0.018 ± 0.015} & 0.344 ± 0.343 & 0.142 ± 0.040 &  0.907 ± 0.715  \\
    &CounterSynth~\cite{pombo2023equitable} & - & 28.24 ± 1.31 & 92.84 ± 0.95 & 0.030 ± 0.018 & \textbf{0.014 ± 0.010} & 0.310 ± 0.311 & 0.127 ± 0.035&0.881 ± 0.672 \\
    \cline{2-10}
    & \multicolumn{9}{l}{\textbf{Sequence-aware (Longitudinal)}} \\
    \cline{2-10}
    &SADM~\cite{yoon2023sadm} & - & 26.94  ± 2.28 & 85.18 ± 2.72 & 0.035 ± 0.027 & 0.018 ± 0.015 & 0.329 ± 0.328 & 0.107 ± 0.028 & 0.924 ± 0.705 \\
    &BrLP~\cite{puglisi2024enhancing} & - & 28.51  ± 1.77  & 91.52 ± 1.31 & {0.024 ± 0.026} & {0.015 ± 0.013} & {0.289 ± 0.271} & {0.093 ± 0.025} & {0.863 ± 0.681}\\
    
    \rowcolor{gray!15} 
    & \ourmodel &     -        & 28.37 ± 1.70 & 92.01 ± 1.29 & {0.029 ± 0.030} & {0.017 ± 0.015} & {0.305 ± 0.298} & {0.103 ± 0.030} & {0.863 ± 0.728}\\ 
            
    \rowcolor{gray!15}    
    & \ourmodel & Spatial   & {29.44 ± 1.18} & {93.23 ± 0.97} & {0.022 ± 0.020} & {0.014 ± 0.013} & {0.233 ± 0.257} & {0.083 ± 0.023} & {0.815 ± 0.644}\\

    \rowcolor{gray!15} 
    & \ourmodel~(Proposed) & Fourier & \textbf{29.72 ± 1.04} & \textbf{93.60 ± 0.96} &  \textbf{0.018 ± 0.014} &{0.014 ± 0.011} & \textbf{0.217 ± 0.242} & \textbf{0.079 ± 0.020} & \textbf{0.799 ± 0.604}\\
    \hline
    \end{tabular}%
    }
\end{table*}

\begin{figure*}
    \centering
    \includegraphics[width=1\linewidth]{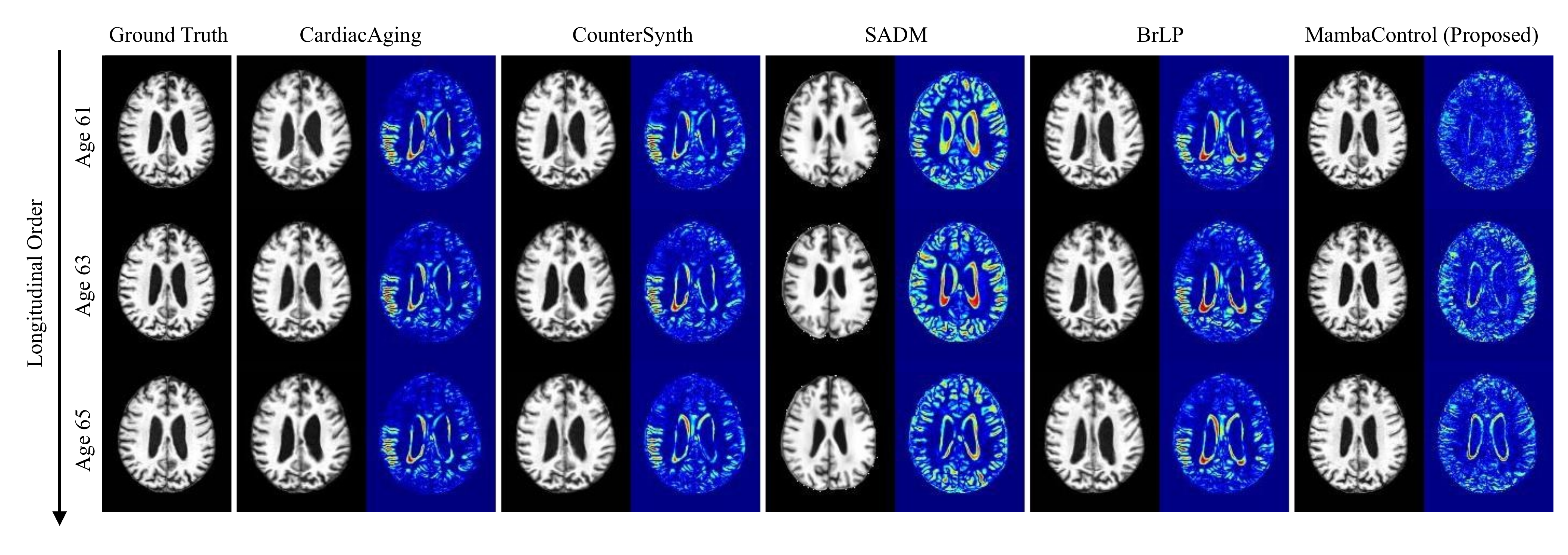}
    \caption{Qualitative evaluation of disease progression models. Each row represents a time point in longitudinal order (top to bottom). For each method, brain MRI predictions (left) are shown with corresponding residual maps (right) showing deviations from Ground Truth.}
    \label{fig:comparison}
\end{figure*}










\section{Experiments and Results}

\subsection{Experimental setup}

\subsubsection{Dataset}
Our study analyzes AD progression through longitudinal imaging analysis. The AD cohort derives multi-phase ADNI 3 dataset~\cite{petersen2010alzheimer}, which provides longitudinal clinical, cognitive, imaging, and biomarker data. All imaging data undergo baseline-aligned rigid registration, with specific preprocessing following established protocols. The inclusion criteria mandate $\geq2$ imaging sessions per subject with complete demographic (age, sex) and clinical metadata. We implement a rigorous 7:1:2 training/validation/test split to ensure robust evaluation.


\subsubsection{Evaluation Metrics}

We evaluated the accuracy of our predictions against ground truth follow-up scans using complementary metrics. At the image level, we assessed overall similarity using the Peak Signal-to-Noise Ratio (PSNR) and the Structural Similarity Index Measure (SSIM). For anatomically specific evaluation, we adopted the BrLP methodology~\cite{puglisi2024enhancing}, calculating the Mean Absolute Error (MAE) between predicted and ground truth voxel intensities within regions vulnerable to AD-related neurodegeneration. These targeted regions included the hippocampus, amygdala, lateral ventricles, thalamus, and cerebrospinal fluid spaces.

\subsubsection{Implementation Details}
We trained our proposed model using AdamW~\cite{loshchilov2017decoupled} (learning rate = $1\mathrm{e}{-4}$) and utilized off-the-shelf VAE encoders for latent features extraction. To ensure stable Mamba training and prevent NaNs, we employed gradient clipping (norm$\leq2.0$) and weight decay (0.01). Mixed-precision training was adopted to reduce computational overhead. All experiments were performed on two NVIDIA V100 (32GB) GPUs.

\subsection{Comparison and Ablation Study}

\subsubsection{Comparison with the state-of-the-arts}
\label{sec:baseline}
The existing state-of-the-arts can broadly be classified into two categories: single-image (cross-sectional) methods and sequence-aware (longitudinal) methods. Single-image models, exemplified by CardiacAging~\cite{campello2022cardiac} and CounterSynth~\cite{pombo2023equitable}, aim to predict disease progression based solely on a single brain MRI scan. In contrast, sequence-aware approaches such as SADM~\cite{yoon2023sadm} and BrLP~\cite{puglisi2024enhancing} incorporate temporal information by utilising a series of historical brain MRIs, thereby capturing longitudinal patterns in the data. For comparative purposes, SADM was re-implemented following the original design, while BrLP was evaluated using its publicly released pre-trained weights. For all methods, prediction of the current MRI is conditioned on the most recent available scan within the input sequence.


Table~\ref{tab:performance} summarizes the performance comparison between our proposed method, MambaControl (including ablation results), and state-of-the-art approaches. In terms of SSIM, the single-image method CounterSynth achieved 92.84\%, surpassing the sequence-aware BrLP (91.52\%) and our MambaControl model without graph integration (92.01\%). However, integrating spatial and Fourier graphs enabled MambaControl to achieve the highest SSIM scores of 93.23\% and 93.60\%, respectively. Regarding PSNR, BrLP (28.51dB) and MambaControl without graph integration (28.37dB) performed slightly better than CounterSynth (28.24dB). Notably, MambaControl with Fourier graph integration attained the top PSNR score of 29.72dB. Finally, for regional volume MAE, while BrLP demonstrated lower error than CounterSynth and MambaControl without graph integration, our full MambaControl model incorporating spatial and Fourier graphs yielded the most accurate regional predictions, surpassing the baseline methods.

A significant advantage of MambaControl is its parameter efficiency. Our model achieves state-of-the-art performance using only 215.5 million generator parameters, significantly fewer than BrLP's 553.2 million, demonstrating superior parameter efficiency.

A qualitative comparison is illustrated in Figure~\ref{fig:comparison}. Diffusion-based methods, such as SADM, BrLP, can exhibit over-smoothed or noticeably deviated outputs when compared to traditional single-image approaches. This blurring effect is also evident in MambaControl when without graph, but mitigated after graph integration.

\subsubsection{Ablation Study}

Our ablation study revealed that baseline MambaControl, while achieving competitive SSIM (92.01\%) and PSNR (28.37dB), showed regional MAE performance that lagged slightly behind BrLP. This deficiency stems from the baseline SSM's lack of spatial inductive biases inherent in UNets. Incorporating the graph-based control mechanism directly addressed this limitation. Adding the spatial graph improved SSIM to 93.23\%, and the Fourier graph further enhanced SSIM to 93.60\% and PSNR to 29.72dB. Most critically, graph integration significantly reduced regional MAE, demonstrating its effectiveness in imparting the necessary anatomical locality guidance absent in the baseline SSM.

\section{Conclusion}

We introduced MambaControl to overcome the prevalent challenges of anatomical inconsistency and poor temporal modeling in disease progression prediction. Our approach uniquely integrates a Mamba state-space model for long-range dependency capture with a ControlNet enhanced by spatial and spectral graph representations to enforce structural constraints. This synergy enables MambaControl to achieve state-of-the-art results, demonstrating superior overall image quality and markedly improved anatomical fidelity. Notably, it significantly reduces volume prediction errors in critical disease regions relative to existing techniques. MambaControl thus offers enhanced reliability and interpretability, constituting a significant step forward in developing tools for personalized prognosis and clinical decision support based on disease trajectory analysis.

\section*{ACKNOWLEDGMENT}
This work was supported by the grant from Science and Technology Development Fund of Macao (0004/2024/E1B1) and the Macao Polytechnic University Grant (RP/FCA-08/2024).

\bibliographystyle{IEEEtran}
\bibliography{myBib}

\end{document}